# CoAtNeXt: An Attention-Enhanced ConvNeXtV2–Transformer Hybrid Model for Gastric Tissue Classification

*Mustafa YURDAKUL[1*]* 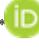, *Şakir TAŞDEMİR[2]*

[1*]Kırıkkale University, Computer Engineering Dept, Kırıkkale, Türkiye, mustafayurdakul@kku.edu.tr
[2]Selçuk University, Computer Engineering Dept, Konya, Türkiye, stasdemir@selcuk.edu.tr

**Abstract:**

**Background and objective**

Early diagnosis of gastric diseases is crucial to prevent fatal outcomes. Although histopathologic examination remains the diagnostic gold standard, it is performed entirely manually, making evaluations labor-intensive and prone to variability among pathologists. Critical findings may be missed, and lack of standard procedures reduces consistency. These limitations highlight the need for automated, reliable, and efficient methods for gastric tissue analysis.

**Methods**

In this study, a novel hybrid model named CoAtNeXt was proposed for the classification of gastric tissue images. The model is built upon the CoAtNet architecture by replacing its MBConv layers with enhanced ConvNeXtV2 blocks. Additionally, the Convolutional Block Attention Module (CBAM) is integrated to improve local feature extraction through channel and spatial attention mechanisms. The architecture was scaled to achieve a balance between computational efficiency and classification performance. CoAtNeXt was evaluated on two publicly available datasets, HMU-GC-HE-30K for eight-class classification and GasHisSDB for binary classification, and was compared against 10 Convolutional Neural Networks (CNNs) and ten Vision Transformer (ViT) models.

**Results**

CoAtNeXt achieved 96.47% accuracy, 96.60% precision, 96.47% recall, 96.45% F1 score, and 99.89% AUC on HMU-GC-HE-30K. On GasHisSDB, it reached 98.29% accuracy, 98.07% precision, 98.41% recall, 98.23% F1 score, and 99.90% AUC. It outperformed all CNN and ViT models tested and surpassed previous studies in the literature.

**Conclusion**

Experimental results show that CoAtNeXt is a robust architecture for histopathological classification of gastric tissue images, providing performance on binary and multiclass. Its highlights its potential to assist pathologists by enhancing diagnostic accuracy and reducing workload.

## 1. Introduction

The gastric tissue is an organ in which a large number of cells with different functions coexist. Its building blocks, such as fat cells, smooth muscle cells, connective tissue elements, mucous membrane, mucin-secreting cells and cells of the immune system, fulfill mechanical, chemical and immune-related functions of the gastric mucosa[1, 2]. While this cellular and textural diversity makes the stomach resilient and flexible physiologically, it also complicates disease assessment. In particular, gastric cancer is a disease characterized by the disruption of its complex textural structure and poses a significant health problem as the fifth most common type of cancer worldwide[3, 4]. Gastric cancer, which is the third leading cause of cancer-related deaths, is usually symptomatic at an advanced stage, making early detection vital[5].

At this point, histopathologic examination stands out as the gold standard method in the diagnosis of gastric cancer[6, 7].

Biopsy samples taken from suspicious areas are colored with special dyes such as Hematoxylin and Eosin (H&E) and evaluated in detail by pathologists under a microscope. Through histopathological examination, not only the presence of the tumor, but also critical details such as the degree of differentiation of tumor cells, their relationship with surrounding tissues, inflammatory response and tissue architecture are determined. Accurate differentiation of different cell and tissue types is crucial for both detecting cancer and determining the most appropriate treatment approach. However, classical microscopic evaluation methods are based on subjective interpretations and require intense attention. There are significant limitations such as the high workload of pathologists, long working hours, the risk of missing critical details during the examination and the difficulty of standardizing results. Therefore, digital pathology and artificial intelligence assisted image analysis technologies are becoming increasingly important. Moreover, advanced artificial intelligence methods enable more accurate and standardized diagnostic support systems[8-13].

Many studies have been conducted in the literature using AI techniques for gastric tissue analysis.

Sharma et al.[14] proposed a novel CNN model with 9 layers for automatic analysis of digital histopathology images for gastric cancer diagnosis. The model consists of three convolutions, three pooling and three fully connected layers. In the study, H&E stained gastric cancer images from 11 patients were used. The proposed model achieved 69.9% accuracy in cancer classification and 81.4% accuracy in necrosis detection.

Li et al. [15] developed a Hierarchical Conditional Random Field-based approach (HCRF-AM) that combines the Attention Mechanism (AM) module with the Image Classification (IC) module. In this method, CNN is trained on selected attention regions and classification is performed. Evaluated on a dataset of 700 gastric histopathology images, the model achieved 91.4% accuracy on the test data.

Wang et al.[16] proposed the CrossU-Net model for segmentation of gastric precancerous lesions (GPL) in gastric cancer using RGB and hyperspectral images. The proposed model achieved 96.53% accuracy and 91.62% Dice score. However, the high dimensionality of hyperspectral data created a significant computational burden.

Ba et al.[17] examined the impact of a DL-assisted method on 16 pathologists in the diagnosis of gastric cancer. In the study, 110 histopathologic images were evaluated in two separate sessions, with and without DL assistance. Significant improvements in diagnostic accuracy (AUC: 91.1% vs. 86.3%), sensitivity (90.63% vs. 82.75%) and examination time (22.68 seconds vs. 26.37 seconds) were obtained when DL support was used. Especially in inexperienced pathologists, this support provided significant benefit.

Lou et al.[18] developed a comprehensive and expert-validated histopathological image dataset named HMU-GC-HE-30K, containing approximately 31,000 patches that classify the tumor microenvironment components of gastric cancer into eight distinct classes. They also demonstrated the classification potential of the dataset by reporting AUC values of 96% and 94% using EfficientNet-B0 and ViT models, respectively.

Hu et al.[19] proposed a novel publicly available histopathology dataset, GasHisSDB, which consists of images of normal and abnormal gastric cancer tissues. In their study, sub-datasets of 80×80, 120×120 and 160×160 pixels were created and the effect of different sizes on classification performance was examined. In deep learning-based experiments, they achieved 96.12% accuracy in 80×80, 96.47% in 120×120 and 95.90% in 160×160 dimensions with the VGG16 model.

Yong et al.[20] evaluated various pretrained CNN architectures on the GasHisSDB dataset, and the most successful result on the 80x80 pixel subset was achieved by the DenseNet169 model, which outperformed the other evaluated CNN architectures by achieving 96.67% accuracy, 96.70% AUC and 96.1% F1-score.

Yong et al.[21] tested different variants of Mobilenet, EfficientNet, DenseNet, Inception and Xception architectures on GasHisSDB dataset. The best 5 models were used with ensemble technique using an unweighted averaging method. The proposed ensemble model achieved 97.72% accuracy, 97.65% AUC, 97.39% precision, 97.18% recall and 97.28% F1-score on the 80×80 pixel sub-database.

Khayatian et al. [22]used the EfficientNetV2B0 model as a feature extractor to classify 80x80 gastric histopathology images in the GasHisSDB dataset and obtained deep features and fed these features to the CatBoost classifier, reporting 89.7% accuracy, 87% precision, 86% recall, 91% specificity and 87% F1-score.

Studies in the literature have developed various methods for automatic classification of gastric histopathologic images. However, these approaches cannot effectively model both local features and global contextual features concurrently. Convolutional architectures perform strongly on local details but have limited learning of global relations, while transformer-based architectures miss local details while modeling long-range relations. Furthermore, many studies have focused only on binary classification, neglecting multi-class gastric tissue discrimination, and reducing the generalization capability of the models. This makes it difficult to provide the comprehensive and reliable classification performance required in clinical diagnosis.

All these limitations require the development of holistic approaches for more accurate and detailed analysis of gastric histopathology images. In this study, we propose a novel DL model that aims to overcome the existing limitations and contribute to both the literature and clinical practice. The main contributions of the study are summarized as follows:

- A novel hybrid model, CoAtNeXt, has been designed in which CBAM-enhanced Improved ConvNeXtV2 blocks are integrated instead of the MBConv layers of the CoAtNet architecture. It combines the strength of convolutional layers for local feature extraction and transformer-based mechanisms for global contextual relations modeling.
- The addition of the CBAM module to ConvNeXtV2 blocks overcomes the challenges of complex morphological structures and high intraclass variation in gastric histopathology images, enabling adaptive focus on regions of interest in channel and spatial dimensions.
- The model architecture, which scales balanced in layer depth and channel width, is designed to limit computational cost while maintaining classification performance.
- Extensive experiments were conducted on two large and publicly available datasets (HMU-GC-HE-30K for multiclass classification and GasHisSDB for binary classification) and comparative analyses were performed with 10 different CNN and 10 ViT architectures. In all experiments, 5-fold cross-validation was applied.
- The proposed CoAtNeXt model provided higher accuracy and reliability in both binary and multiclass tasks compared to existing methods and studies in the literature, demonstrating its potential to contribute to clinical histopathological image analysis.

The rest of the paper is organized as follows: Section 2 describes the material and methods. Section 3 explains the experimental setup and evaluation metrics. Section 4 presents the experimental results, Section 5 provides the discussion, and the final section concludes the study.

**2. Material and methods**
**2.1. Datasets**
All DL models evaluated in the study were tested on two different large datasets. The HMU-GC-HE-30K dataset was used to test multi-class classification with gastric tissue images that are divided into eight classes, while the GasHisSDB dataset was used for binary classification between abnormal and normal tissues.

**2.1.1. HMU-GC-HE-30K**
HMU-GC-HE-30K is a publicly available histopathologic image dataset of gastric tissues. It was created by digitizing 300 slides from Harbin Medical University Cancer Hospital. Each slide was divided into 224×224 pixel patches, resulting in 31,096 images. All images were labeled by three pathologists into eight tissue types: adipose tissue (ADI), necrotic debris (DEB), lymphocyte accumulation (LYM), mucus (MUC), muscle tissue (MUS), normal mucosa (NOR), connective tissue (STR), and tumor tissue (TUM), with 3,887 images per category. ADI is fat tissue used to assess invasion depth. DEB is cell debris indicating regression or high cell death. LYM shows dense lymphocyte clusters or immune infiltration. MUC indicates mucous secretions important for diagnosing mucinous adenocarcinomas. MUS is the stomach's muscle layer relevant for invasion depth. NOR is healthy mucosa used as a reference. STR is fibrous connective tissue that provides support and may be involved in invasion. TUM is malignant epithelial tissue with nuclear atypia and high mitotic activity. Sample images for each category are shown in Fig. 1.

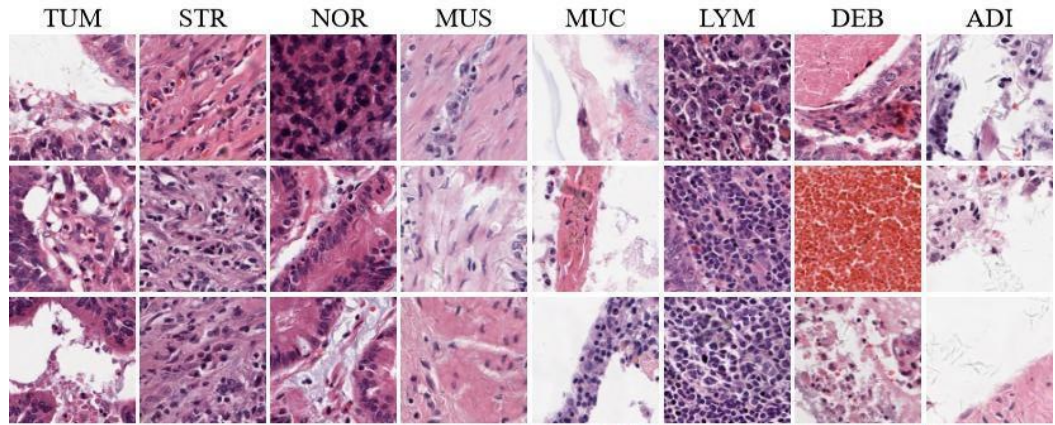

**Fig. 1.** Sample images from the HMU-GC-HE-30K dataset, illustrating eight histopathological gastric tissue classes.

### 2.1.2. GasHisSDB

GasHisSDB is a publicly available dataset consisting of labelled microscopic images prepared for the histopathological diagnosis of gastric cancer. The dataset was obtained from 600 original images of 2048×2048 size sourced from Shanghai University of Traditional Chinese Medicine. A total of 245,196 small-sized (sub-size) patches were created by cropping cancer areas marked by expert pathologists.

The dataset is divided into three sub-datasets containing images of 160×160, 120×120, and 80×80 pixels. In this study, the 80x80 pixel sub-set was preferred, as it preserves the broader tissue context and allows for the analysis of micromorphological features. This sub-set consists of a total of 146.651 labelled images, including 59.151 abnormal and 87.500 normal images. Normal images consist of regular, single-layered cell arrangements without cancer cells. Abnormal images, on the other hand, contain cancer cells with irregular arrangements and distinct nuclear atypia. Sample images for each category are shown in Fig. 2.

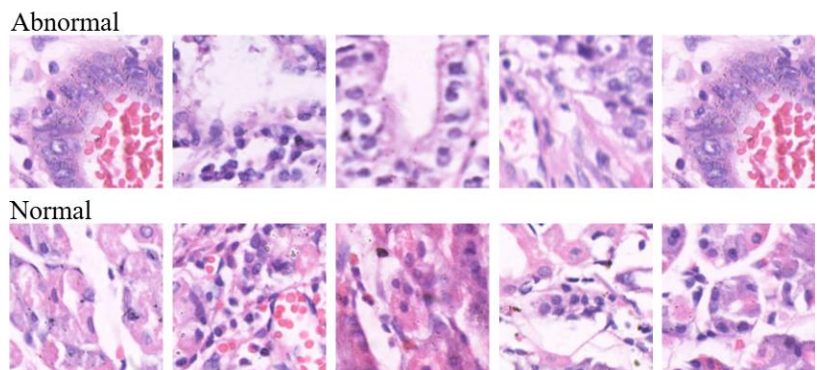

**Fig. 2.** Sample images from the GasHisSDB dataset, showing normal and abnormal gastric tissues.

### 2.2. CoAtNet

CoAtNet[23] is a DL architecture that combines convolutional structures and transformer mechanisms.

The model consists of five stages from S0 to S4, and the final section includes global average pooling (GAP) and fully connected (FC) layers to perform the classification task. In the S0 layer, two $3 \times 3$ convolution blocks are applied to the input image to perform basic feature extraction. S0 stage detects low-level patterns such as edges and corners. In the S1 and S2 layers, local features are learned using MBConv blocks. These blocks have a strong feature extraction capacity despite their efficiency-focused structure. In the S3 and S4 stages of the model, relative positional (ReIPos) transformer blocks are used. In these blocks, attention calculations are performed based on fixed biases related to both the similarity between input vectors and the relative distance between positions. CoAtNet's relative attention mechanism is defined as in Eq. 1.

$$y_i = \sum_{j \in G} \frac{exp\ (x_i^T x_j + b_{i-j})}{exp\ (x_i^T x_k + b_{i-k})} \cdot x_j \tag{1}$$

In Eq.1. $x_i$ and $x_j$ are vector representations of different locations at the input; $b_{i-j}$ is a constant bias based on the relative distance between locations; $G$ is the global location space. With relative attention, the model is able to compute both input-sensitive and location-sensitive attention, thus modeling the global context more effectively. CoAtNet has a flexible architecture that can be easily adapted for different tasks and resource constraints. With different variants ranging from CoAtNet-0 to CoAtNet-5, the model scales in terms of number of layers ($L$) and channel width ($D$). In this way, structures that can address a wide range of solutions are obtained, from applications requiring low computational cost to high-accuracy scenarios. In this study, CoAtNet-0 was preferred as it provides a good balance between accuracy and computational efficiency, making it suitable for large-scale histopathology image classification tasks with limited resources.

## 2.3. CBAM

CBAM [24] is an attention mechanism for both spatial and channel attention. It produces a discriminative feature map by progressively weighting the input feature map in the channel and spatial dimensions. In the first stage, the Channel Attention Module (CAM) performs average and maximum pooling operations on the input feature map, generating two descriptors that are passed through shared MLP layers, summed, and activated via a sigmoid function to produce the channel attention map $M_C$ as seen Eq. 2.

$$M_C = \sigma(MLP(AvgPool) + MLP(MaxPool)) \tag{2}$$

This feature map is multiplied by the input feature to produce an intermediate feature map $F'$ which is rescaled according to the channel size importance. Then the Spatial Attention Module (SAM) performs average and maximum pooling on $F'$ in channel size. These two maps are concatenated, passed over a convolution layer and a spatial attention map $M_S$ is generated by applying sigmoid activation as seen Eq. 3.

$$M_S = \sigma(Conv([AvgPool(F'); MaxPool(F')])) \tag{3}$$

In the final stage, $F'$ and $M_S$ are multiplied element-wise to obtain a refined feature $R_F$ map focused in both channel and spatial dimensions. With this structure, CBAM achieves adaptive emphasis on important channels and spatial regions through sequential channel and spatial attention.

## 2.4. ConvNeXtV2

ConvNeXt [25] is a backbone module proposed as an alternative to transformer-based models by redesigning classic convolutional neural networks with modern DL techniques.

In this module, large-kernel depthwise convolution layers enhance feature extraction by capturing broader contextual information. In addition, Layer Normalization and GELU activation are integrated to regulate gradient flow and improve training stability. ConvNeXtV2[26] builds on this architecture and specifically addresses the problem of feature collapse during training.

For this purpose, the Global Response Normalization (GRN) layer is introduced, which computes the L2 norm of each channel's spatial map as defined in Eq. 4 and balances the activation magnitudes to produce more consistent and distinguishable feature representations.

$$G(X)_i = \left\| X_i \right\|_2 \tag{4}$$

Here, $Xi$ denotes the i-th channel of the input feature map. The computed values are subsequently normalized according to Eq. 5 to model the relative importance among channels.

$$N(G(X)_i) = \frac{N(X)_i}{\sum_{j=1}^{C} GX_j} \tag{5}$$

Finally, the original feature map is scaled by normalization factor and combined with a residual connection, as shown in Eq. 6.

$$\widehat{X_i} = \gamma . X_i . N(G(X)_i) + \beta + X_i \tag{6}$$

Here, γ and β are learnable parameters. This process prevents channels from learning overly similar representations by providing a channel competition mechanism, enabling more discriminative and rich features to be obtained. The ConvNeXtV2 block performs spatial feature extraction using depthwise convolution, provides balanced activation transformations with normalization and GELU layers, and increases channel-level selectivity with a GRN layer in the final stage. However, the original ConvNeXtV2 block relies solely on channel-based normalization and has limited attention capability for identifying important regions in complex visual data. To overcome this, we propose the Improved ConvNeXtV2 by integrating the CBAM immediately after the normalization layer. CBAM applies sequential channel and spatial attention, helping the model better focus on both what and where is important. This enhancement emphasizes relevant regions while suppressing noise, aiming to improve classification performance, particularly in detailed datasets such as histopathological images. Structural diagrams of ConvNeXtV2 and Improved ConvNeXtV2 are shown in Fig. 3.

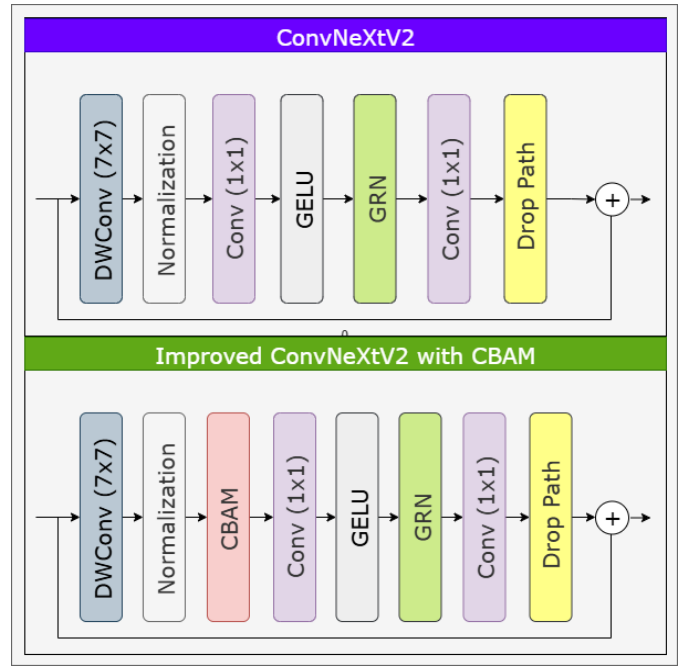

Fig. 3. Structural diagram comparing the original ConvNeXtV2 block and the proposed Improved ConvNeXtV2 block with integrated CBAM for enhanced channel and spatial attention.

## 2.5. CoAtNeXt

CoAtNeXt is a hybrid model for gastric tissue classification based on the CoAtNet architecture. CoAtNet has an architecture consisting of MBConv layers and ReIPos transformer blocks. However, while MBConv layers are lightweight and computationally efficient, their feature-extraction capacity is limited and may be insufficient to capture complex morphological details, especially in histopathological images. To overcome this limitation, the CoAtNeXt integrates CBAM-enhanced Improved ConvNeXtV2 blocks—each containing channel and spatial attention mechanisms and internal residual connections—instead of the conventional MBConv layers in the S1 and S2 stages of CoAtNet. Thus, the model is able to better discriminate important textural information while enhancing local feature extraction.

In stages S3 and S4, the ability to model long-range contextual relationships is maintained by retaining the transformer blocks with ReIPos encoding; each transformer block also uses a residual skip-connection to stabilize gradient flow. Between every two stages, a 1×1 linear projection together with stride-2 down-sampling both halves the spatial dimensions and doubles the channel depth—transitioning feature maps to ensure smooth resolution reduction and capacity growth. This hybrid architecture thus learns robust local features in early ConvNeXtV2 stages while preserving powerful global context modeling in deeper transformer stages. Fig. 4 shows the schematic overview of the proposed CoAtNeXt architecture.

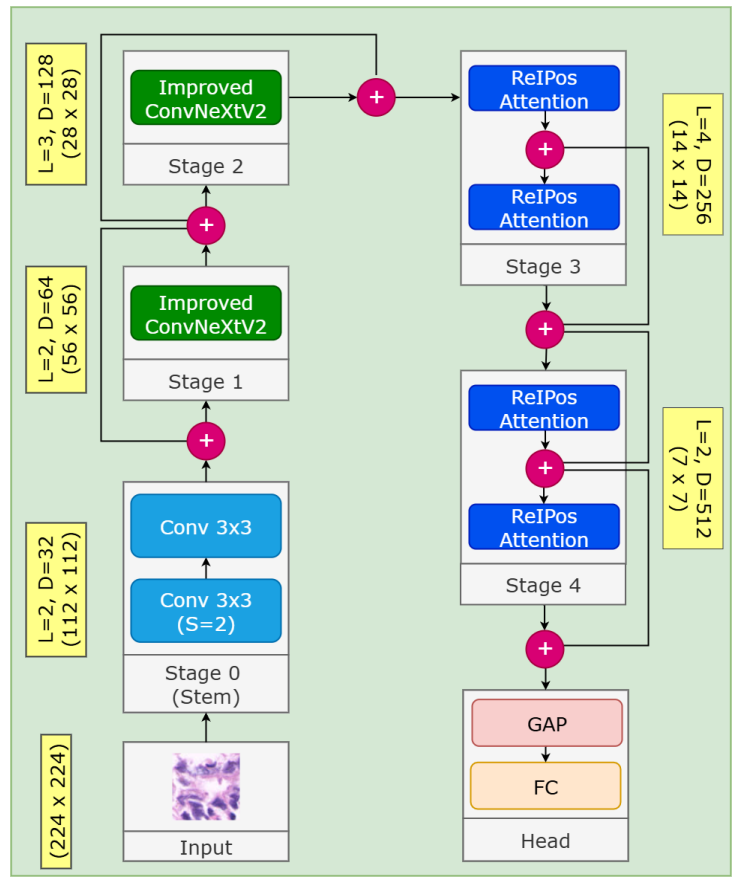

**Fig. 4.** Schematic overview of the proposed CoAtNeXt architecture, showing stage-wise integration of Improved ConvNeXtV2 blocks and transformer layers for hybrid local-global feature extraction

Table 1. Detailed stage-wise architectural configuration of the proposed CoAtNeXt model, including block types, layer counts (L), and channel dimensions (D).

| Stage | Block Type | L | D |
| --- | --- | --- | --- |
| Stage 0-Conv | Conv3×3 | 2 | 32 |
| Stage 1 | Improved ConvNeXtV2 | 2 | 64 |
| Stage 2 | Improved ConvNextV2 | 3 | 128 |
| Stage 3-TFMRel | Rel. Pos. Transformer | 4 | 256 |
| Stage 4-TFMRel | Rel. Pos. Transformer | 2 | 512 |
| Head | GlobalAvgPool + FC | – | - |

The model architecture has been precisely structured to balance the computational load and the performance introduced by the addition of the Improved ConvNeXtV2 blocks. Architectural details are provided in Table 1. The number of layers (L) and channel sizes (D) of the model are planned to gradually increase the representation capacity while balancing computational cost. In particular, increasing the number of channels to 64 and 128 in S1 and S2 and to 256 and 512 in S3 and S4 ensures that the model can learn more complex and discriminative features as it deepens. While early stages focus on extracting low-level edge and texture patterns, deeper stages build high-dimensional contextual representations. The increased channel width also provides the rich feature vectors required for transformer blocks to work effectively, reducing information loss and gradually enhancing information transfer while maintaining parameter efficiency.

Finally, the Head—consisting of global average pooling followed by a fully-connected classification layer—produces either eight-class or binary outputs, depending on the dataset used.

## 3. Implement details

All models tested in this study were developed using Python programming language and trained using TensorFlow 2.14.0 and Keras 2.11.4 libraries. The experiments were performed on a Windows 11 computer with 128 GB RAM and 2 x Nvidia RT3090 (24 GB) GPUs. GPU acceleration was achieved using CUDA v12.7. For model training, categorical cross entropy loss function and Stochastic Gradient Descent (SGD) optimization algorithm were used. Batch size was set as 32 and the number of epochs was set as 50. All hyperparameters were selected with the grid search method. During the training and evaluation process, 5-fold cross-validation was applied to analyze the generalization ability of the model.

### 3.1. Evaluation Metrics

To objectively evaluate the performance of models, accuracy, precision, recall, F1-score, and Cohen's kappa metrics, which are commonly used in classification problems, were used. These metrics are based on the TP (true positive), TN (true negative), FP (false positive), and FN (false negative) values obtained from the confusion matrix. While the accuracy metric represents overall success, the precision metric indicates the true positive prediction rate, and the recall metric indicates the true positive capture rate, the F1-score provides a balanced average of these two values. The Kappa metric, on the other hand, provides a fairer assessment by comparing the model's performance to random guessing. AUC (Area Under the Curve) calculates the classification performance at all threshold values by analysing the relationship between the classifier's TPR (True Positive Rate) and FPR (False Positive Rate). The mathematical formulas for the evaluation metrics, accuracy, precision, recall, F1-score and AUC are presented in Eq. 7–11, respectively.

$$Accuracy = (TP + TN)/(TP + TN + FP + FN) \qquad (7)$$

$$Precision = TP/(TP + FP) \qquad (8)$$

$$Recall = TP/(TP + FN) \tag{9}$$

$$F1 - Score = 2(Precision \; x \; Recall)/(Precision + Recall) \tag{10}$$

$$AUC = \int_{0}^{1} TPR(FPR)d(FPR) \tag{11}$$

## 4. Results

All the experiments conducted in the study were carried out using a 5-fold cross-validation method to obtain reliable results. The calculated performance metrics consist of the average values of these folds and are also reported to evaluate the consistency of the performance.

### 4.1. CNN Results

The experimental results of CNN models on the HMU-GC-HE-30K dataset are presented in Table 2. When the table is analyzed, it is seen that the accuracy rates of CNN models vary between 68% and 93%. In addition, the fact that all metrics have close and consistent values indicate that the classification performance of the model is balanced and that it can generalize successfully in different classes without showing overfitting tendency.

When the classification results of models are analyzed, it is seen that classical CNN architectures such as DenseNet121, InceptionV3, MobileNet, Xception and ResNet50 generally had lower accuracy and consistency levels in the range of 68%-72%. In contrast, more modern architectures such as EdgeNeXt, GhostNetV2, ConvNeXtV2, and InceptionNeXt stand out with higher accuracy in the 91%-92% range and generally more balanced metric values. Such a difference shows the impact of advanced connection structures, deeper feature extraction layers and optimization techniques used in the architectural design on classification performance.

The most successful model, CoAtNet, stands out with 93.12% accuracy, 93.33% precision, 93.12% recall, 93.11% F1-score and 99.15% AUC. The high success of CoAtNet can be attributed to its hybrid structure that combines convolution-based local feature extraction and transformer-based attention mechanisms with global contextual relations. In this way, it effectively learns both fine texture details and long-range contextual patterns, and has demonstrated superior performance in the histopathologic image classification task.

**Table 2.** Performance metrics of CNN architectures evaluated on the HMU-GC-HE-30K

| Model | Performance measurement metrics(%) | | | | |
|---|---|---|---|---|---|
| | **Accuracy** | **Precision** | **Recall** | **F1-Score** | **AUC** |
| DenseNet121 [27] | 71.54 | 73.11 | 71.54 | 71.27 | 95.68 |
| InceptionV3 [28] | 71.41 | 72.52 | 71.41 | 71.55 | 95.26 |
| MobileNet [29] | 70.13 | 72.66 | 70.13 | 70.34 | 95.47 |
| **CoAtNet [23]** | **93.12** | **93.33** | **93.12** | **93.11** | **99.15** |
| ResNet50 [30] | 68.79 | 70.69 | 68.79 | 68.80 | 94.83 |
| GhostNetV2 [31] | 92.79 | 92.95 | 92.79 | 92.82 | 99.07 |
| Xception [32] | 72.78 | 73.39 | 72.78 | 72.91 | 95.52 |
| EdgeNeXt | 91.19 | 91.58 | 91.19 | 91.21 | 99.14 |
| [33] | | | | | |
| ConvNeXtV2 [26] | 92.52 | 92.84 | 92.52 | 92.54 | 99.18 |
| InceptionNeXt [34] | 92.30 | 92.43 | 92.30 | 92.31 | 99.15 |

When the performances of CNN algorithms on the GasHisSDB dataset are analyzed, it is seen that high accuracy rates between 90% and 96% are obtained. When Table 3 is analyzed, it is seen that almost all models achieve accuracy rates between 94% and 96%. However, the DenseNet121 model performed lower than the other models with an accuracy of 90.75%. This underperformance is due to the limited parametric capacity of the model, which is prone to overlearning on some datasets.

On the other hand, ConvNeXtV2 was the most successful model in terms of all metrics with 96.73% accuracy, 96.58% precision, 96.62% recall, 96.60% F1 score and 99.49% AUC. This superior performance of the ConvNeXt architecture is achieved thanks to its state-of-the-art feature extractor structure, large-kernel filters and GRN technique.

**Table 3.** Performance metrics of the CNN architectures evaluated on the GasHisSDB dataset

| **Model** | **Performance measurement metrics(%)** | | | | |
|---|---|---|---|---|---|
| | **Accuracy** | **Precision** | **Recall** | **F1-Score** | **AUC** |
| DenseNet121 [27] | 90.75 | 92.09 | 90.33 | 90.12 | 98.75 |
| InceptionV3 [28] | 94.02 | 94.67 | 93.51 | 93.66 | 99.18 |
| MobileNet [29] | 94.92 | 95.15 | 94.48 | 94.69 | 99.26 |
| CoAtNet [23] | 96.66 | 96.58 | 96.49 | 96.53 | 99.51 |
| ResNet50 [30] | 95.15 | 95.10 | 94.84 | 94.95 | 99.04 |
| GhostNetV2 [31] | 95.62 | 95.24 | 95.81 | 95.48 | 99.20 |
| Xception [32] | 95.98 | 96.08 | 95.58 | 95.81 | 99.36 |
| EdgeNeXt [33] | 94.70 | 94.55 | 94.77 | 94.58 | 98.98 |
| **ConvNeXtV2 [26]** | **96.73** | **96.58** | **96.62** | **96.60** | **99.49** |
| InceptionNeXt [34] | 95.23 | 94.99 | 95.39 | 95.09 | 99.34 |

In conclusion, CNN-based models generally perform well in both multiple and binary gastro histopathologic image classification tasks. Particularly new generation architectures exhibit higher accuracy and consistency compared to conventional models. While the accuracy ranges between 68% and 93% in multiple classification, it reaches 90% to 96% in binary classification. Among the most successful models, ConvNeXtV2 extracts local features very effectively thanks to its advanced convolutional structure, while CoAtNet stands out by learning both local and global context relations with its hybrid architecture combining convolution and transformer structures.

Based on these results, the main motivation for this work is the idea of augmenting the convolutional layers in the early stages of CoAtNet with the advanced local feature extractor blocks of ConvNeXtV2. The combination of these two powerful architectures could allow the development of a new generation of hybrid models that are more balanced and high-performance and could provide a significant advance in the classification of gastrohistopathologic images.

### 4.2. ViT results

The experimental results of ViT-based models on the HMU-GC-HE-30K dataset are presented in Table 4. The table shows that the accuracy ranges from approximately 76% to 93%.

The results show that some models such as MaxViT, EfficientViT, PVTV2 and MobileViTV2 show lower accuracy and consistency levels, which may be due to differences in architectural depth, efficiency of attention layers and parameter optimization. In contrast, advanced models such as GPViT, DaViT and SwinTransformerV2 stand out with high accuracy and balanced metric values in the 92%-93% range. This difference demonstrates the contribution of multi-layered attention mechanisms, advanced positional encoding strategies and extensive contextual feature extraction to classification success.

The most successful model, SwinTransformerV2, stands out with 93.19% accuracy, 93.27% precision, 93.19% recall, 93.17% F1-score and 99.11% AUC. This success can be attributed to its ability to effectively learn both local details and long-range contextual relationships using multi-layered and hierarchical attention mechanisms.

**Table 4.** Performance metrics of ViT architectures evaluated on the HMU-GC-HE-30K

| Model | Performance measurement metrics(%) | | | | |
|---|---|---|---|---|---|
| | Accuracy | Precision | Recall | F1-Score | AUC |
| GPViT [35] | 92.83 | 93.08 | 92.83 | 92.86 | 99.24 |
| DaViT [36] | 93.13 | 93.29 | 93.13 | 93.13 | 99.15 |
| EfficientViT [37] | 79.14 | 80.50 | 79.14 | 78.90 | 97.75 |
| GCViT [38] | 86.09 | 86.32 | 86.09 | 86.00 | 98.64 |
| MobileViTV2 [39] | 82.85 | 84.81 | 82.85 | 83.01 | 98.25 |
| **SwinTransformerV2 [40]** | **93.19** | **93.27** | **93.19** | **93.17** | **99.11** |
| TinyViT [41] | 86.52 | 87.40 | 86.52 | 86.43 | 98.60 |
| MaxViT [42] | 75.93 | 78.19 | 75.93 | 75.69 | 97.31 |
| PVTV2 [43] | 80.35 | 80.83 | 80.35 | 80.25 | 97.24 |
| FlexiViT [44] | 85.07 | 85.40 | 85.07 | 84.99 | 98.40 |

The performance of CNN models on the GasHisSDB dataset is presented in Table 5. When the Accuracy metric is analyzed, it is seen that the accuracy rates of the models vary between 83% and 97%.

The GCViT and EfficientViT models performed lower than the other models with 83.40% and 84.99% accuracy rates respectively. This performance can be explained by the fact that these models architecturally focus more on global features and cannot learn local texture details well enough. Local feature extraction is especially important for learning fine texture differences in histopathologic images. On the other hand, PVTv2 was the most successful model in all metrics with 97.58% accuracy, 96.67% sensitivity, 97.35% sensitivity, 97.01% F1 score and 99.80% AUC. PVTv2's performance is based on its more efficient structural design, which is more optimized than pure ViT architectures. In particular, its multilevel structure and progressive scaling approach enables it to process both local and global features together, which allows the model to integrate information at different resolution levels more effectively. Furthermore, PVTv2's more efficient token reduction mechanisms minimize the loss of detail, which is a significant advantage for high-resolution and detailed data such as histopathological images.

**Table 5.** Performance metrics of ViT architectures evaluated on the GasHisSDB

| Model | Performance measurement metrics(%) | | | | |
|---|---|---|---|---|---|
| | Accuracy | Precision | Recall | F1-Score | AUC |
| GPViT [35] | 96.84 | 96.74 | 96.71 | 96.72 | 99.57 |
| DaViT [36] | 96.24 | 96.08 | 96.25 | 96.14 | 99.10 |
| EfficientViT [37] | 84.99 | 85.71 | 85.20 | 84.59 | 93.96 |
| GCViT [38] | 83.40 | 85.20 | 85.50 | 83.26 | 96.32 |
| MobileViTV2 [39] | 94.76 | 94.52 | 94.78 | 94.57 | 99.04 |
| SwinTransformerV2 [40] | 93.11 | 93.58 | 93.92 | 93.74 | 99.12 |
| TinyViT [41] | 96.53 | 96.47 | 96.33 | 96.38 | 99.38 |
| MaxViT [42] | 94.02 | 94.67 | 93.51 | 93.66 | 99.18 |
| **PVTV2 [43]** | **97.58** | **96.67** | **97.35** | **97.01** | **99.60** |
| FlexiViT [44] | 94.83 | 94.73 | 94.59 | 94.62 | 98.99 |

When ViT models are compared with CNN-based models, it is seen that they perform in similar accuracy ranges. On the HMU-GC-HE-30K dataset, CNN models achieved an accuracy between 68% and 93%, while ViT models achieved an accuracy between 76% and 93%. On the GasHisSDB dataset, CNN models show accuracy ranging between 90% and 96%, while the accuracy of ViT-based models ranges between 83% and 97%.

One of the most important findings that can be drawn from these results is that CNN-based models exhibit strong classification performance thanks to their success in local feature extraction, while ViT architectures can achieve similar accuracy levels with their ability to model the global context. All these findings show that a hybrid approach combining the strong features of CNN and ViT architectures has the potential to achieve more robust and generalizable results in gastrohistopathological tissue classification tasks.

**4.3. Ablation Study and Proposed Model results**

The ablation study on the HMU-GC-HE-30K and GasHisSDB dataset was performed to examine the contribution of ConvNeXtV2-based blocks and different attention mechanisms (SE, ECA, CBAM) integrated into these blocks to the classification performance instead of the MBConv layers in the CoAtNet model.

The experimental results on the HMU-GC-HE-30K are presented in Table 6.

CoAtNet, used as the base model, achieved 93.12% accuracy, 93.33% precision, 93.10% recall, 93.11% F1-score and 99.15% AUC. In the first step of the ablation experiment, the MBConv layers in the CoAtNet architecture were directly replaced with standard ConvNeXtV2 blocks (without the addition of any attention mechanism). This version achieved 94.50% accuracy and 99.81% AUC, demonstrating that the backbone change alone provides a significant performance improvement. In the second phase, different attention mechanisms were integrated into ConvNeXt blocks, respectively. When the SE (Squeeze-and-Excitation) module was added, accuracy increased to 94.75% and AUC to 99.85%. When the ECA (Efficient Channel Attention) module was integrated, performance improved even more, reaching 95.64% accuracy and 99.89% AUC. The most successful result was obtained with CBAM integration. Since CBAM includes both channel and spatial attention components, it maximizes the model's capacity to focus on important regions and features. The final model developed with CBAM showed the highest performance with 96.47% accuracy, 96.60% precision, 96.47% recall, 96.45% F1-score and 99.89% AUC. In summary, the proposed CoAtNeXt architecture achieved about +3.35% accuracy improvement and more than +0.74 points AUC improvement compared to the baseline CoAtNet model.

**Table 6.** Ablation study results on the HMU-GC-HE-30K dataset showing the impact of different attention mechanisms integrated into the ConvNeXtV2 blocks.

| Model | Performance measurement metrics(%) | | | | |
|---|---|---|---|---|---|
| | Accuracy | Precision | Recall | F1-Score | AUC |
| CoAtNet | 93.12 | 93.33 | 93.10 | 93.12 | 99.15 |
| CoAtNeXt (ConvNeXtV2) | 94.50 | 94.55 | 94.51 | 94.50 | 99.81 |
| CoAtNeXt (ConvNeXt+SE) | 94.75 | 94.88 | 94.77 | 94.75 | 99.85 |
| CoAtNeXt (ConvNeXt+ECA) | 95.64 | 95.79 | 95.64 | 95.64 | 99.89 |
| **CoAtNeXt (ConvNeXt+CBAM)** | **96.47** | **96.60** | **96.47** | **96.45** | **99.89** |

Similarly, the results of the ablation study on the GasHisSDB dataset are presented in Table 7. The baseline CoAtNet model achieved 96.66% accuracy, 96.58% precision, 96.49% sensitivity, 96.53% F1-score and 99.51% AUC. The model created by replacing the MBConv layers with ConvNeXtV2 blocks showed a small performance improvement with 96.73% accuracy and 99.52% AUC. The addition of the SE attention mechanism improved accuracy to 97.46% and AUC to 99.69%. The integration of the ECA module further improved the accuracy to 97.68% and the AUC to 99.78%. The highest performance was achieved with the CBAM method. The CBAM-enhanced CoAtNeXt model showed the best results with 98.29% accuracy, 98.07% precision, 98.41% sensitivity, 98.23% F1-score and 99.90% AUC.

**Table 7.** Ablation study results on the GasHisSDB dataset showing the impact of different attention mechanisms integrated into the ConvNeXtV2 blocks.

| Model | Performance measurement metrics(%) | | | | |
|---|---|---|---|---|---|
| | Accuracy | Precision | Recall | F1-Score | AUC |
| CoAtNet | 96.66 | 96.58 | 96.49 | 96.53 | 99.51 |
| CoAtNeXt (ConvNeXtV2) | 96.73 | 96.58 | 96.62 | 96.60 | 99.52 |
| CoAtNeXt (ConvNeXtV2+SE) | 97.46 | 97.69 | 97.04 | 97.34 | 99.69 |
| CoAtNeXt (ConvNeXt+ECA) | 97.68 | 97.66 | 97.52 | 97.59 | 99.78 |
| CoAtNeXt (ConvNeXt+CBAM) | **98.29** | **98.07** | **98.41** | **98.23** | **99.90** |

In summary, these results show that integrating ConvNeXtV2 blocks into the CoAtNet model and incorporating different attentional mechanisms (especially CBAM) into these blocks significantly improved classification performance on both datasets. These architectural enhancements strengthened the model's ability to focus on important regions, resulting in higher accuracy.

**4.4. Comparison with literature studies**

The HMU-GC-HE-30K dataset was publicly released in 2025 and there is only one experimental study on this dataset. Lou et al. reported an AUC of 96% with the EfficientNet-B0 architecture. While EfficientNet-B0 is successful in local feature extraction, it may have limitations in modeling the global context, which may have led to this level of performance. Furthermore, in Lou et al.'s study, only the AUC metric was used for evaluation. In contrast, the proposed CoAtNeXt model achieved 96.47% accuracy and 99.89% AUC on the HMU-GC-HE-30K dataset, a significant improvement. The comparative values are presented in Table 8.

**Table 8.** Comparison of the proposed model's performance with existing literature results on the HMU-GC-HE-30K dataset

| Study | Performance measurement metrics(%) | | | | |
|---|---|---|---|---|---|
| | Accuracy | Precision | Recall | F1-Score | AUC |
| **Lou et al.** | - | - | - | - | 96 |
| **This study** | **96.47** | **96.60** | **96.47** | **96.45** | **99.89** |

On the other hand, various studies have been conducted on the GasHisSDB dataset in the literature. In this context, EfficientNet by Khyantin et al., VGG16 by Hu et al., and DenseNet169 as well as CNN-based ensembles by Yong et al. were employed. In Table 9, a comparison of the performance values of the proposed method and those from the literature is presented. It is clearly seen that the proposed method achieves the highest classification performance.

The main reason for this success is that the CNN-based models used in the approaches in the literature are not able to adequately capture the fine details in histopathologic images and the global features representing the distinctive microscopic structures.

**Table 9.** Comparison of the proposed model's performance with existing literature results on the GasHisSDB

| Study | Performance measurement metrics(%) | | | | |
| --- | --- | --- | --- | --- | --- |
| | Accuracy | Precision | Recall | F1-Score | AUC |
| Khayatian et al. | 89.7 | 87 | 86 | 87 | - |
| Hu et al. | 96.12 | Abnormal: 94.2<br>Normal: 97.4 | Abnormal: 96.3<br>Normal: 96 | Abnormal: 95.2<br>Normal: 96.7 | - |
| Yong et al. | 96.67 | - | - | 96.1 | 96.70 |
| Yong et al. | 97.72 | 97.39 | 97.18 | 97.28 | 97.65 |
| **This study** | **98.29** | **98.07** | **98.41** | **98.23** | **99.90** |

In summary, the CoAtNeXt model, with its hybrid architecture combining convolutional and transformer mechanisms, was able to efficiently extract both locally and globally significant features, making it the most successful method in the literature on both datasets.

## 5. Discussion

Artificial intelligence-based gastric tissue classification is critical for the early diagnosis of gastric diseases and provides support to specialists in the diagnostic process. In this study, experimental analyses were performed on two large open-access datasets. In the first experiment, a total of 10 CNN models were tested on the HMU-GC-HE-30K and GasHisSDB datasets. The CoAtNet model achieved a 92.13% accuracy rate on the HMU-GC-HE-30K dataset, while the ConvNeXtV2 model achieved a 96.73% accuracy rate on the GasHisSDB dataset, making them the most successful CNN models.

In the second stage, a total of 10 different ViT models were evaluated on the same data sets. SwinTransformerV2 achieved the best results with an accuracy rate of 93.19% on HMU-GC-HE-30K, while PVTV2 achieved the best results with an accuracy rate of 97.58% on GasHisSDB.

In the third stage, the performance of the proposed hybrid CoAtNeXt model was evaluated. The CoAtNeXt model demonstrated superior performance with accuracy rates of 96.47% on HMU-GC-HE-30K and 98.29% on GasHisSDB. When compared to the 10 CNN and 10 ViT models tested and existing studies in the literature, the performance of the proposed CoAtNeXt model yielded the most successful results on both datasets.

As a result of the experimental studies, confusion matrices were created to see the class-level performance of the most successful CNN, ViT and the proposed CoAtNeXt models.

Fig. 5 shows the confusion matrices of the CoAtNet, SwinTransformerV2 and CoAtNeXt models on the HMU-GC-HE-30K dataset. The CoAtNet model performs well overall, but has higher false positive and false negative values in the MUC, STR and TUM classes. SwinTransformerV2 improved the overall correct classification rate, but errors remained significant in the DEB and MUC classes. The CoAtNeXt model stands out with higher correct classification rates and lower misclassification values in all classes. The density on the matrix diagonal shows that the model is able to separate all classes more successfully and balanced.

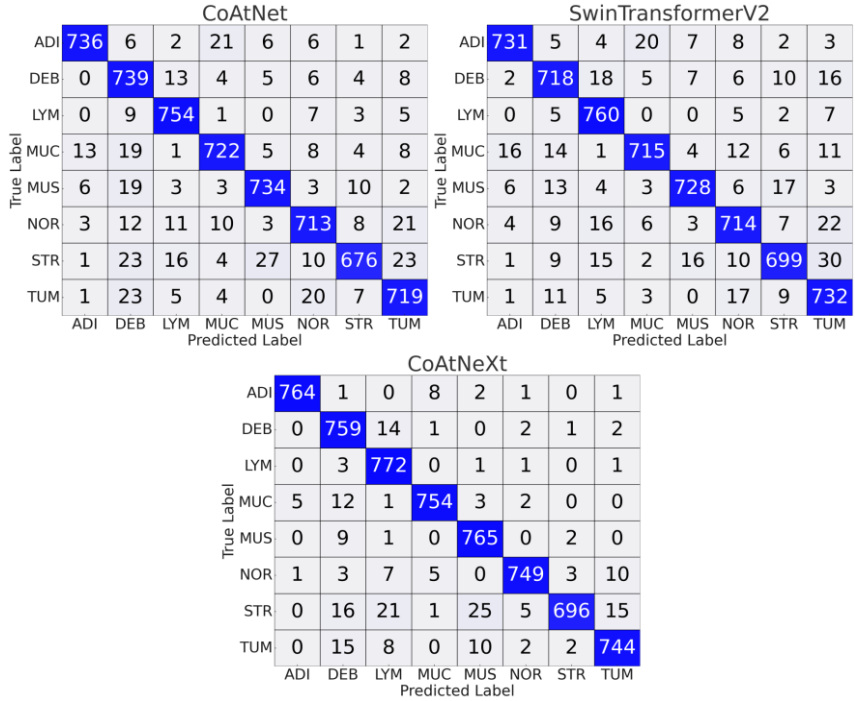

**Fig. 5.** Confusion matrices comparing CoAtNet, SwinTransformerV2, and CoAtNeXt models on HMU-GC-HE-30K

Fig. 6 shows the confusion matrices of the ConvNeXtV2, PVTv2 and CoAtNeXt models on the GasHisSDB dataset. Although the ConvNeXtV2 model performs well overall, its false positive and false negative values are higher than the other models. Although the PVTv2 model partially improved the correct classification of abnormal samples, the number of false negatives is still high. In contrast, the proposed CoAtNeXt model achieved the highest correct classification rates and the lowest error values in both classes.

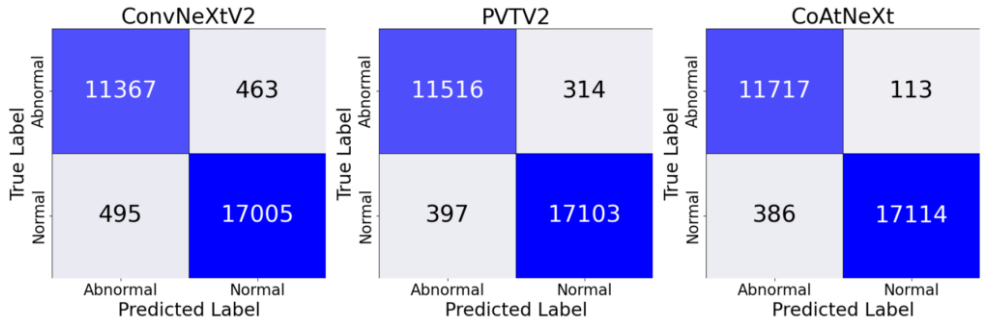

**Fig. 6.** Confusion matrices comparing ConvNeXtV2, PVTV2, and CoAtNeXt models on GasHisSDB.

The number of parameters and model size of deep learning models are of great importance in terms of both training time and hardware requirements in real-time applications. Therefore, Table 10 shows a comparison of the number of parameters and model sizes of the most successful CNN and ViT-based models and the proposed CoAtNeXt architecture.

**Table 10.** Model complexity comparison in terms of parameter count and model size.

| Model | Parameter Count(M) | Size (MB) |
|---|---|---|
| CoAtNet | 25 | 319 |
| SwinTransformerV2 | 49.7 | 222 |
| ConvNeXtV2 | 28.6 | 130 |
| PVTV2 | 62.6 | 241 |
| CoAtNeXt | 18.8 | 110 |

With only 18.8 million parameters and a size of 110 MB, the CoAtNeXt model has a much lighter structure than models such as SwinTransformerV2 (49.7M / 222 MB) and PVTv2 (62.6M / 241 MB). Despite this, its higher classification performance demonstrates that the model offers a balanced structure in terms of both computational efficiency and accuracy, thereby increasing its potential for use in clinical applications.

This study also has limitations. The histopathological images used were prepared using only the H&E staining technique. The performance of the model has not yet been evaluated on images obtained using other histopathological staining methods. The lack of open-access datasets for different staining techniques in the literature makes this evaluation difficult. In this context, future studies will be able to overcome this limitation by collaborating with healthcare institutions to create a new dataset consisting of histopathological images prepared using different staining methods.

## 6. Conclusion

In this study, CoAtNeXt, a hybrid model enhanced with attention mechanisms for the histopathological classification of gastric tissue images, is proposed. The proposed model is developed based on the CoAtNet architecture and integrates ConvNeXtV2 blocks supported by the CBAM module instead of classic MBConv layers. As a result, the model has been able to effectively extract local features and successfully model long-range contextual relationships. In experiments conducted on large-scale, open-access datasets such as HMU-GC-HE-30K and GasHisSDB, the CoAtNeXt model outperformed all tested CNN and ViT models, demonstrating superior classification performance. On the HMU-GC-HE-30K dataset, it achieved 96.47% accuracy, 96.60% precision, 96.47% sensitivity, 96.45% F1 score, and 99.89% AUC; In the GasHisSDB dataset, high values such as 98.29% accuracy, 98.07% precision, 98.41% sensitivity, 98.23% F1 score, and 99.90% AUC were obtained.

When compared to existing studies in the literature, the CoAtNeXt model was found to provide more reliable and accurate results in both binary and multi-class classification tasks. Additionally, it provides a highly efficient structure in terms of the number of parameters used (18.8 million) and model size (110 MB), which is a significant advantage for its use in clinical applications.

**CRediT authorship contribution statement**
Mustafa Yurdakul and Şakir Taşdemir conducted the entire conceptualization, experimentation, analysis, interpretation, writing, and revision of the study collaboratively.

**Declaration of competing interest**
The author declares no competing interests.

**Data availability**
HMU-GC-HE-30K is available at the following link.
https://figshare.com/articles/dataset/Gastric_Cancer_Histopathology_Tissue_Image_Dataset_GCHTID_/25954813
GasHisSDB is available at the following link.
https://figshare.com/articles/dataset/GasHisSDB/15066147/1?file=28969725

**Declaration of generative AI in the writing process**
During the preparation of this manuscript, the author used GPT-4 to enhance the readability and language of the text. The author thoroughly reviewed and edited all content to ensure accuracy and integrity, and takes full responsibility for the published work.